\RequirePackage{fix-cm}
\documentclass[twocolumn]{svjour3}          %

\smartqed  %
\usepackage{hyperref}
\hypersetup{
    colorlinks=true,
    breaklinks=true,
    linkcolor=red,
    filecolor=magenta,      
    urlcolor=blue,
    citecolor=blue
}
\usepackage{pifont}
\usepackage{multirow}
\usepackage{amsmath}
\usepackage{rotating}
\usepackage[dvipsnames]{xcolor}
\usepackage{comment}
\usepackage{amssymb}

\usepackage{array} %

\usepackage[ruled,vlined]{algorithm2e}
\usepackage[T1]{fontenc}
\usepackage{lmodern}
\usepackage{tabularx}
\usepackage{mathtools}
\usepackage[caption=false]{subfig}
\usepackage{booktabs}
\usepackage{amsfonts}

\usepackage{pgfplots}
\usepgfplotslibrary{groupplots}
\pgfplotsset{compat=1.18}

\usepackage{graphicx}
\usepackage{stfloats} 
\usepackage{placeins}   
\newcommand{\cmark}{\ding{51}}  
\newcommand{\best}[1]{\textbf{#1}}
\newcommand{\secondbest}[1]{\underline{#1}}

\begin{document}
\sloppy
\journalname{International Journal on Document Analysis and Recognition (IJDAR)}

\title{TableSeq: Unified Generation of Structure, Content, and Layout}

\author{Laziz Hamdi \and Amine Tamasna \and Pascal Boisson \and Thierry Paquet}

\institute{
L. Hamdi \and T. Paquet \at
LITIS, Rouen, Normandy, France \\
\email{laziz.hamdi@univ-rouen.fr} \\
\email{thierry.paquet@univ-rouen.fr}
\and
A. Tamasna \and P. Boisson \at
Malakoff Humanis, Paris, France \\
\email{amine.tamasna@malakoffhumanis.com} \\
\email{pascal.boisson@malakoffhumanis.com}
}

\date{Received: date / Accepted: date}

\maketitle

\begin{abstract}
We present \emph{TableSeq}, an image-only, end-to-end framework for joint table structure recognition, content recognition, and cell localization. The model formulates these tasks as a single sequence-generation problem: one decoder produces an interleaved stream of \texttt{HTML} tags, cell text, and discretized coordinate tokens, thereby aligning logical structure, textual content, and cell geometry within a unified autoregressive sequence. This design avoids external OCR, auxiliary decoders, and complex multi-stage post-processing. TableSeq combines a lightweight high-resolution FCN-H16 encoder with a minimal structure-prior head and a single-layer transformer encoder, yielding a compact architecture that remains effective on challenging layouts. Across standard benchmarks, TableSeq achieves competitive or state-of-the-art results while preserving architectural simplicity. It reaches 95.23 TEDS / 96.83 S-TEDS on PubTabNet, 97.45 TEDS / 98.69 S-TEDS on FinTabNet, and 99.79 / 99.54 / 99.66 precision / recall / F1 on SciTSR under the CAR protocol, while remaining competitive on PubTables-1M under GriTS. Beyond TSR/TCR, the same sequence interface generalizes to index-based table querying without task-specific heads, achieving the best IRDR score and competitive ICDR/ICR performance. We also study multi-token prediction for faster blockwise decoding and show that it reduces inference latency with only limited accuracy degradation. Overall, TableSeq provides a practical and reproducible single-stream baseline for unified table recognition, and the source code will be made publicly available at \url{https://github.com/hamdilaziz/TableSeq}.


\end{abstract}

\section{Introduction}
\label{sec:intro}
\begin{figure}[t]
\centering
\includegraphics[width=0.99\linewidth]{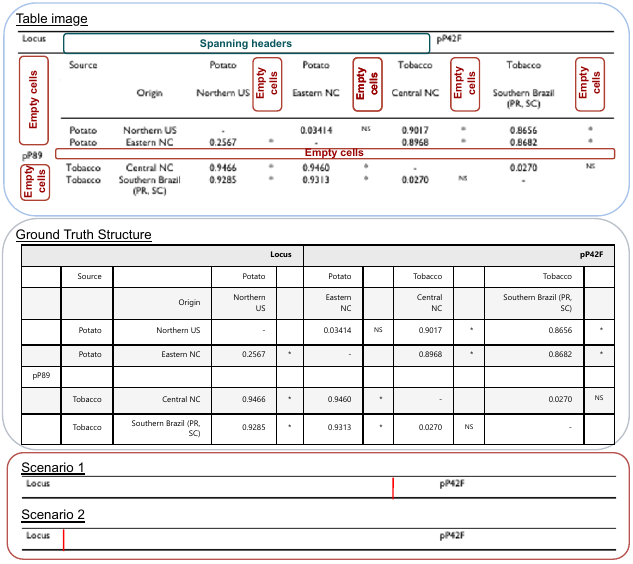}
\caption{PubTabNet sample “PMC2048936\_004\_00” from the training set, with no column separators, few line separators, and misaligned headers, which make it difficult to interpret the global structure; multiple interpretations are plausible.}
\label{fig:complex_sample}
\end{figure}

Tables serve as essential vehicles for encoding structured information across scientific publications, financial reports, legal documents, and technical manuals, yet the automated conversion of table images into semantically rich, machine-readable markup remains a formidable challenge in document intelligence. This task requires the simultaneous recovery of three interdependent elements: (1) logical structure (row/column hierarchies, multi-span headers, merged cells), (2) physical layout (precise cell bounding boxes), and (3) cell content (machine-readable text or symbols). Moreover, this task must also cope with complex real-world scenarios including borderless tables (see Fig.~\ref{fig:complex_sample}) with nested headers, rotated orientations, heterogeneous domain shifts, and inconsistent annotation standards that complicate both model training and benchmark evaluation. Recent comprehensive surveys~\cite{YU2024128154,huang2024detection} highlight that despite significant progress, performance gaps persist in handling visually complex tables from domains like biomedical literature or engineering schematics, where structural irregularities and domain-specific formatting conventions exacerbate recognition difficulties.

Historically, approaches to table recognition have evolved significantly. Early methods often relied on heuristic, rule-based pipelines that used traditional computer vision techniques to detect lines and contours to infer the grid structure. While effective for simple, clearly demarcated tables, these methods proved brittle and failed to generalize to the more complex, "in-the-wild" tables that are very common today. With the advent of deep learning, the field shifted towards more robust, data-driven models. The initial wave of these models treated TSR as a component-based problem. Some focused on object detection, training models to identify key components like cells and then using complex post-processing or graph-based reasoning to reconstruct the table's relational structure \cite{Schreiber2017DeepDeSRTDL,paliwal2019tablenet}. Subsequent research further refined this graph-based paradigm to better model the complex relationships between cells \cite{Xue2021TGRNet,Qiao2021LGPMA,Liu2022NCGM,Raja2020TopDown}. An alternative but related line of work pursued a split-and-merge strategy. These methods first predict the locations of row and column separators to define a grid of basic cells, which are then merged to form multi-span cells \cite{li2020tablebank}. More recent transformer-based designs falling within this paradigm, such as TRUST and TSRFormer, have shown improved robustness, particularly for unlined and rotated tables \cite{guo2208trust,lin2022tsrformer}.

More recently, the field has seen a paradigm shift towards end-to-end image-to-sequence (Im2Seq) models, which formulate TSR as a generation task. This approach was popularized by PubTabNet~\cite{zhong2020image}, which introduced a large-scale dataset and proposed generating the table's HTML markup directly from the image using an encoder-dual-decoder architecture. Additionally, this work introduced the Tree-Edit-Distance-based Similarity (TEDS) metric, which has since become the de facto standard for evaluating TSR performance. Following this breakthrough, subsequent research has explored various architectural and methodological refinements. TableFormer adapted the DETR framework to jointly predict structure tokens and cell bounding boxes using transformer decoders \cite{nassar2022tableformer,smock2023aligning}. Other works, such as TableMaster~\cite{Ye2021TableMaster}, have demonstrated state-of-the-art performance by leveraging powerful pre-trained backbones and sophisticated decoder designs. Indeed, a key challenge in Im2Seq models is to link the logical structure generated (e.g., a \texttt{<td>} tag) with its corresponding physical location in the image. To address this, various strategies have been proposed. Some models improve decoding efficiency by generating a more compact structure language instead of verbose HTML \cite{lysak2023optimized}. Others have developed novel mechanisms to tighten this logical-physical coupling: VAST~\cite{huang2023improving} uses a dedicated coordinate-sequence decoder, LORE~\cite{xing2023lore} frames the problem as logical-location regression to bind structure tokens to grid positions, and TFLOP~\cite{khang2025tflop} introduced a layout-pointer mechanism to directly associate the generated tags with image regions without heuristic matching.

\begin{figure*}[!t]
\centering
\includegraphics[width=\linewidth]{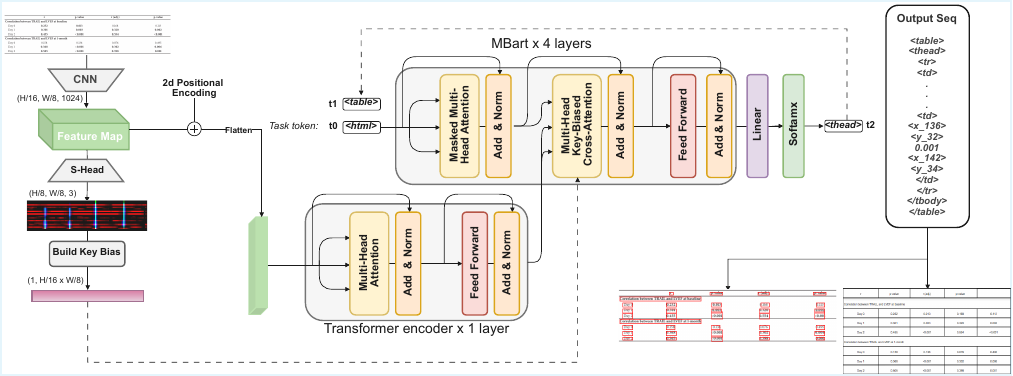}
\caption{Overall framework. The input image is encoded by a lightweight CNN and a single-layer Transformer encoder. A structure-prior head predicts row, column, and corner cues, which are used to bias decoder cross-attention. The decoder then autoregressively generates a single sequence containing HTML tags, cell content, and discretized cell coordinates.}
\label{framework}
\end{figure*}

Recent Im2Seq advances commonly depend on complex pipelines (e.g., multi-decoder setups, task-specific heads, pointer modules), which can hinder reproducibility and deployment. We take the opposite path and prioritize simplicity: a lightweight CNN feature extractor, a single transformer encoder layer, and a standard BART decoder jointly produce \emph{one} interleaved sequence of \texttt{HTML} structure tags, cell text, and discretized bounding-box coordinates, aligning structure–content–geometry by construction and avoiding post-processing. Our core contributions are as follows:
\begin{itemize}
  \item Unified Im2Seq architecture: A minimalist encoder–decoder for end-to-end TSR/TCR (Fig.~\ref{framework}) that emits structure, content, and locations in a single stream without auxiliary decoders or heuristics.
  \item Training recipe for complex layouts: A practical training regimen with lightweight structure supervision, a one-layer transformer, and a simple synthetic-to-real curriculum that improves robustness to spans, nested headers, and grouped columns.
  \item Faster decoding via MTP: A multi-token prediction variant that reduces autoregressive iterations and wall-clock latency with small accuracy trade-offs.
  \item Comprehensive ablations: A systematic study quantifying the effect of key choices, including encoder type and vertical resolution ($H'/32$ vs.\ $H'/16$), structure supervision and key-biasing, grid quantization for coordinates, use of synthetic data. We report TEDS gains separately on \emph{simple} vs.\ \emph{complex} tables and include cross-dataset transfer results to clarify where each component helps.
\end{itemize}

\section{Related Work}
\label{sec:related_work}

\subsection{Bottom-Up Methods}
Bottom-up approaches dominated early deep learning solutions by detecting primitive components (cells, text blocks) and reconstructing structural relationships through a post-processing stage. These methods typically employ object detection or segmentation frameworks to identify table elements before inferring row/column hierarchies. LGPMA introduced local and global pyramid mask alignment to improve cell boundary precision through multiscale context aggregation \cite{Qiao2021LGPMA}. NCGM addressed nested headers via neural collaborative graph machines that match cell graphs to structural templates \cite{Liu2022NCGM}. TGRNet proposed hierarchical graph parsing for complex table reconstruction using graph neural networks \cite{Xue2021TGRNet}, while DeepDeSRT pioneered graph-based reasoning where cells serve as nodes and edges represent row/column relationships \cite{Schreiber2017DeepDeSRTDL}. TableNet combined segmentation and classification to predict column separators and cell regions \cite{paliwal2019tablenet}, later extended by CascadeTabNet's end-to-end cascade framework for joint table detection and structure recognition \cite{Prasad2020CascadeTabNet}. GTE (Global Table Extractor) leveraged visual context for simultaneous table identification and recognition of cell structure through a region proposal network \cite{zheng2021global}. Recent advances include TRACE's corner-edge alignment for robust geometric reconstruction \cite{baek2023trace} and dynamic query-based detectors that improve structural reasoning through DETR architectures \cite{Raja2020TopDown}. Though effective for bordered tables, these methods struggle with borderless tables and nested structures due to error propagation in multi-stage pipelines.

\subsection{Split-and-Merge Based Methods}
This paradigm first predicts row/column separators to form a base grid of atomic cells, then merges them into multi-span structures. SEMv2 advanced separation line detection using conditional convolution that adapts to heterogeneous document layouts \cite{zhang2023semv2}. Deep Splitting and Merging explicitly modeled hierarchical decomposition through recursive splitting operations followed by merge decisions \cite{tensmeyer2019deep}, while Split, Embed and Merge (SEME) introduced embedding-based similarity metrics for robust cell merging \cite{zhang2022split}. TRUST achieved state-of-the-art results through hierarchical transformer modeling that handles rotated and borderless tables by progressively refining structural hypotheses \cite{guo2208trust}. TSRFormer extended this with transformer-based feature refinement and adaptive merging strategies for complex industrial documents \cite{lin2022tsrformer}. Visual Understanding of Complex Tables incorporated segmentation collaboration to resolve ambiguities in nested headers \cite{raja2022visual}, and Scene Table Recognition improved alignment through cross-modal feature fusion for document images with heterogeneous backgrounds \cite{wang2023scene}. These methods demonstrate superior robustness for irregular layouts but suffer from misalignment between logical structure and physical coordinates when intermediate representations are imperfect.

\subsection{Image-to-Sequence Methods}
Im2Seq models generate structured markup from images through end-to-end sequence modeling, eliminating multi-stage pipelines. PubTabNet pioneered this approach with an encoder-dual-decoder architecture that generates HTML markup and introduced the TEDS evaluation metric \cite{zhong2020image}. TableFormer adapted DETR for joint prediction of structure tokens and bounding boxes through a transformer decoder \cite{nassar2022tableformer}, while TableVLM leveraged multi-modal pre-training to enhance cross-modal alignment between visual and structural features \cite{chen2023tablevlm}. VAST strengthens the link between the predicted table logic and its page geometry by interleaving box coordinates with structural tokens in the decoder’s output stream, ensuring each cell is immediately grounded to its visual location \cite{huang2023improving}. LORE reframed TSR as logical-location regression to bind structure tokens to grid positions \cite{xing2023lore}, and TFLOP introduced a layout-pointer mechanism that associates tags with image regions without heuristic matching \cite{khang2025tflop}. OTSL optimized efficiency through compact tokenization that reduces sequence length by 62\% compared to HTML \cite{lysak2023optimized}. Recent works include ReS2TIM's syntactic structure reconstruction from table images \cite{xue2019res2tim}, Parsing Table Structures in the Wild's robust decoding for diverse document types \cite{long2021parsing}, and OmniParser's unified framework for simultaneous text spotting, KIE, and table recognition \cite{wan2024omniparser}. These methods achieve state-of-the-art performance but many rely on complex architectures with multiple decoders or specialized heads, creating deployment challenges in resource-constrained environments.
Many recent TSR systems achieve strong performance by introducing additional architectural components, such as separate decoders for structure and content prediction, external OCR modules coupled with heuristic post-processing, or explicit alignment mechanisms for linking logical tokens to image regions. While effective, these design choices can increase architectural complexity and make end-to-end deployment less straightforward. In contrast, we adopt a unified single-stream formulation that jointly generates logical structure (\texttt{HTML} tags), textual content, and physical coordinates within one autoregressive sequence, thereby reducing multi-stage dependencies while preserving logical--physical alignment through discrete spatial tokenization.

\section{Methodology}
\label{method}

\subsection{Encoder}
Our encoder is a lightweight CNN (Fig.~\ref{framework}) adapted from the efficient design of Coquenet et al.~\cite{coquenet2022end}. To better preserve fine table cues, we keep the horizontal stride at $W/8$ and \emph{increase} the vertical resolution to $H/16$ (versus $H/32$ in the original), which our ablations show is critical on document tables. We also replace the nonlinearity with \emph{SiLU}~\cite{elfwing2018sigmoid}.

\paragraph{Structure-prior head}
To inject layout cues into the sequence model, we attach a shallow \emph{structure-prior head} that predicts three geometric primitives: row separators, column separators, and cell corners directly from the backbone features. The head (Conv\,{+}\,BN\,{+}\,ConvT\,{+}\,SiLU\,{+}\,Conv) produces a three-channel probability map at a vertically upsampled resolution
\[
\hat{\mathbf{S}}\in[0,1]^{3\times (H/8)\times (W/8)}.
\]
We upsample to $(H/8)$ along $H$ because thinner separators are otherwise poorly represented on a sizeable fraction of images. This head is trained with an auxiliary loss, and its predictions are \emph{also used at inference} to build a cross-attention bias (Sec.~\ref{sec:decoder}).

\paragraph{Sequence encoding}
Let $\mathbf{F}\in\mathbb{R}^{C\times H'\times W'}$ be the backbone feature map with $H'=H/16$, $W'=W/8$, and $C{=}1024$. Before flattening, we apply 2D Rotary Positional Embeddings (RoPE)~\cite{su2024roformer} to retain spatial information in the resulting sequence. We then obtain
\[
\mathbf{S}\in\mathbb{R}^{T_k\times d},\qquad T_k=H'W',\ \ d=1024,
\]
and pass $\mathbf{S}$ through a \emph{single-layer} Transformer encoder (4 heads; model width $d=1024$), which captures long-range structure at a low computational cost. The resulting memory is exploited by the decoder via cross-attention.

\subsection{Decoder}
\label{sec:decoder}
We adopt the first four layers of mBART~\cite{lewis2019bart} and replace the vanilla cross-attention with \emph{Multi-Head Key-Biased Cross-Attention}. For queries $\mathbf{Q}\in\mathbb{R}^{T_q\times d}$, keys $\mathbf{K}\in\mathbb{R}^{T_k\times d}$, values $\mathbf{V}\in\mathbb{R}^{T_k\times d}$, and an additive mask $\mathbf{M}\in\mathbb{R}^{1\times T_q\times T_k}$, we augment the logits with a \emph{key-only} bias derived from the structure prior:
\begin{equation}    
\boxed{\ \mathrm{Attn}_{\text{KB}}(\mathbf{Q},\mathbf{K},\mathbf{V})=\mathrm{softmax}\!\Big(\tfrac{\mathbf{Q}\mathbf{K}^\top}{\sqrt{d}}+\mathbf{M}+\mathbf{B}\Big)\mathbf{V} }
\end{equation}

where $\mathbf{b}\in\mathbb{R}^{T_k}$ is a per-key logit offset and $\mathbf{B}=\mathrm{broadcast}(\mathbf{b})\in\mathbb{R}^{1\times 1\times T_k}$ is shared across heads and query positions. The bias is computed from $\hat{\mathbf{S}}$ by resizing its three maps to $(H',W')$, forming axis profiles for rows and columns, and combining them with the corner confidence to emphasize keys that lie near plausible separators and cell junctions; the final $\mathbf{b}$ is standardized, scaled, and clamped for numerical stability. This design concentrates attention on structurally relevant memory slots without altering the decoder architecture.

\subsection{Tokenization}
To supervise structure explicitly, the target sequence includes \texttt{HTML} tags widely used for table recognition (e.g., \verb|<thead>|, \verb|<tbody>|, \verb|<tr>|, \verb|<td>|) together with a task selector \verb|<html>|. To analyze what the model learns for span tokens, Fig.~\ref{fig:span-embeddings} shows UMAP and PCA projections of their embeddings. The two token families \verb|colspan_k| and \verb|rowspan_k| form well-separated clusters with smooth trajectories as $k$ increases, suggesting that the model encodes both span type and span extent in a structured manner. Inspired by Pix2Seq~\cite{chen2021pix2seq} and its extensions~\cite{chen2022pix2seqv2,kil2023units,huang2024detection}, we represent cell bounding boxes using \emph{discrete location tokens}. Specifically, we introduce
\[
\{\texttt{<x\_0>},\ldots,\texttt{<x\_{999}>}\}\quad\text{and}\quad \{\texttt{<y\_0>},\ldots,\texttt{<y\_{999}>}\},
\]
which encode absolute $x$- and $y$-coordinates after quantization. Continuous coordinates are mapped to this vocabulary on a fixed grid where one unit corresponds to $5$\,px, enabling consistent, resolution-agnostic supervision.
\begin{figure}[t]
  \centering
  \includegraphics[width=\linewidth]{images/spans_clusters.pdf}
  \caption{Embeddings of span tokens projected with UMAP (left) and PCA (right). \texttt{colspan\_k} (green circles) and \texttt{rowspan\_k} (red squares) form distinct clusters with smooth trajectories as $k$ increases; shaded ellipses indicate 95\% confidence regions. In the original embedding space, the two families are well separated.}
  \label{fig:span-embeddings}
\end{figure}
\subsection{Training Objective}
The model autoregressively generates an interleaved sequence of structural (\texttt{HTML}), textual, and location tokens. We optimize the token-level cross-entropy
\begin{equation}
    \mathcal{L}_{\text{seq}}=\mathrm{CE}(\hat{\mathbf{y}},\mathbf{y}),
\end{equation}
and train the structure-prior head with a combination of binary cross-entropy and Dice loss,
\begin{equation}
\mathcal{L}_{\text{prior}}=\mathrm{BCE}+0.5\,\mathrm{Dice}.
\end{equation}
The total objective is
\begin{equation}    
\mathcal{L}=\mathcal{L}_{\text{seq}}+\mathcal{L}_{\text{prior}}.
\end{equation}
During training, gradients through the inference-time bias path are detached: the structure head is supervised solely by $\mathcal{L}_{\text{prior}}$. We use teacher forcing and inject light noise (3\%) for robustness: textual tokens can be substituted by nearby alternatives under a character-error-based confusion model, and location tokens are randomly perturbed by $\pm 3$ indices on each axis.

\subsection{Synthetic Data Generation}
\label{sec:synthetic}

\begin{figure}[t]
  \centering
  \includegraphics[width=\columnwidth]{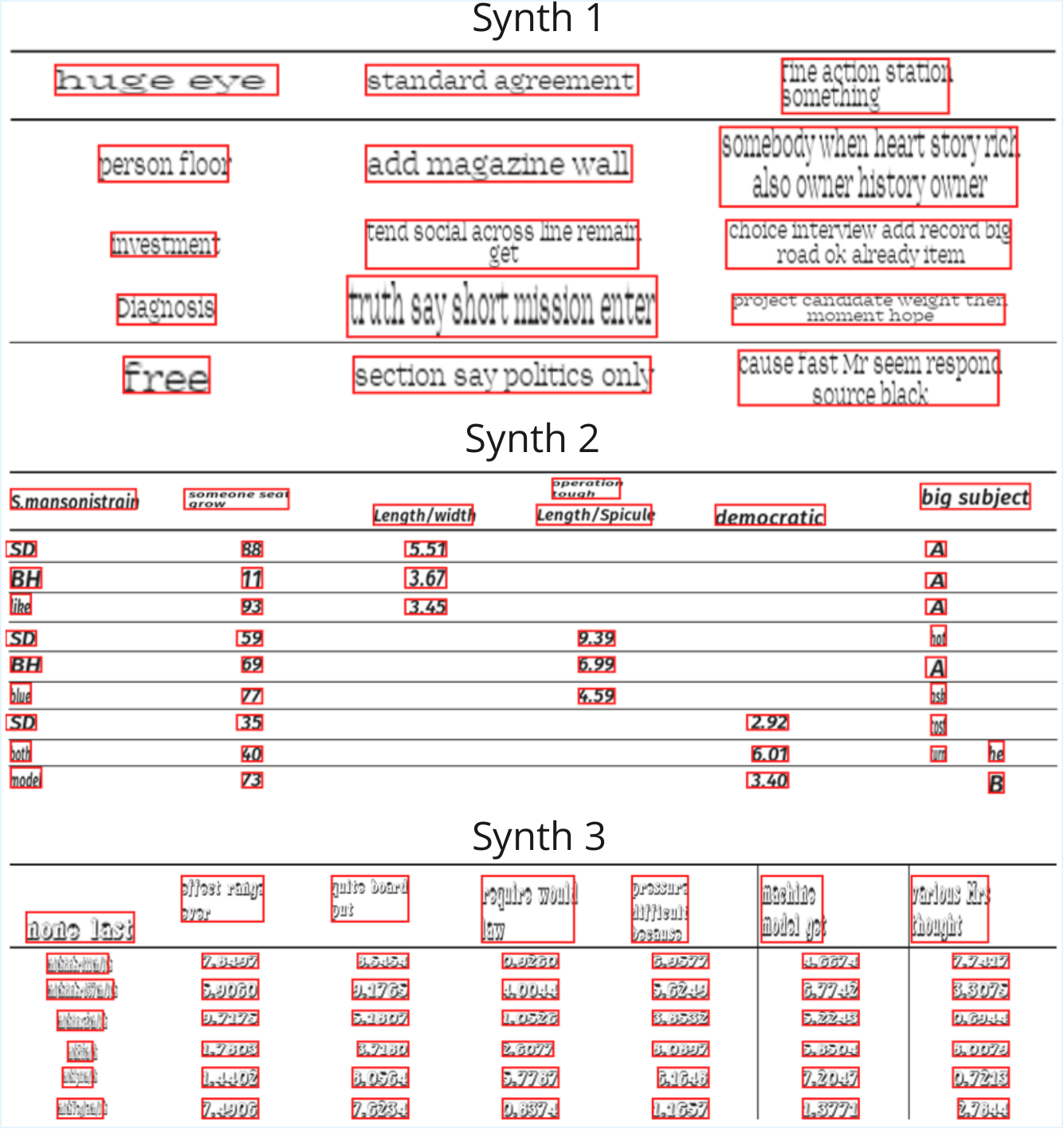}
  \caption{Synthetic page edits produced by our in-place pipeline with cell content boxes; originals from PubTabNet.}
  \label{fig:synth_samples}
\end{figure}

We synthesize paired image--\texttt{HTML} examples (see Fig.~\ref{fig:synth_samples}) by \emph{editing real tables in place}. Starting from an original page image and its \texttt{HTML} with per-cell coordinates, we (i) optionally perturb the \emph{structure} to diversify layouts, and (ii) replace each cell’s text with newly rendered content that is fitted to the cell geometry while preserving the page’s rulings and background. This keeps structure and boxes faithful to the (possibly augmented) layout, while diversifying content, typography, and formatting.
Algorithmic details of the in-place editing procedure are provided in Appendix~\ref{app:synthetic_data}.

\paragraph{Notes on structure augmentation}
We apply \emph{structure-aware} edits stochastically to avoid merely retyping content: span insertion/expansion (and the inverse), shallow header nesting, column grouping labels, and limited row/column insertion/deletion near existing content. Each operation adjusts \texttt{rowspan}/\texttt{colspan}, cell indices, and (if needed) box coordinates so that the HTML remains valid and rectangular. We keep perturbations small to preserve realism and to maintain compatibility with the original ruling lines. These edits can be combined with light photometric jitter on $\tilde I$ (e.g., contrast/brightness noise) if desired.

\paragraph{Outputs}
The pipeline returns $(\tilde I,\tilde H)$ where (i) rulings and backgrounds are inherited from the original page, (ii) every cell contains newly rendered text with fitted glyph geometry, and (iii) the \texttt{HTML} inner content of each \texttt{<td>} is replaced by the escaped text bracketed by normalized coordinates. This preserves logical structure and geometry while diversifying content, fonts, alignments, spacing, and local color/contrast.

\paragraph{Data Normalization and Image Preprocessing}
\label{sec:normalization}

All images are normalized channel-wise using the empirical mean and standard deviation computed on the \emph{training} split. Given an RGB image $I\in[0,255]^{H\times W\times 3}$, we first scale to $[0,1]$ and apply
\[
\tilde I(c) \;=\; \frac{I(c)/255 - \mu_c}{\sigma_c}\qquad\text{for } c\in\{R,G,B\},
\]
where $(\mu_c,\sigma_c)$ are the per-channel statistics aggregated over training images only. The same $(\mu,\sigma)$ are used for validation and test to avoid leakage.

\paragraph{Quality improvement for PubTabNet}
For \textbf{PubTabNet}, we apply a lightweight document-enhancement pass before normalization to reduce background shading and improve local contrast while preserving fine rulings. The routine follows four steps: (1) illumination correction via grayscale morphological opening to flatten background; (2) CLAHE on luminance (LAB) to boost local contrast; (3) mild unsharp masking to sharpen glyph edges; and (4) optional non-local means denoising (kept small to avoid erasing thin lines). Defaults match the code: \texttt{clahe\_clip}=$2.0$, \texttt{clahe\_tiles}=$(8,8)$, \texttt{unsharp\_amount}=$0.5$, \texttt{unsharp\_sigma}=$1.0$, and \texttt{denoise\_h}=\texttt{None}.

\begin{algorithm}[t]
\small
\DontPrintSemicolon
\KwIn{BGR image $I_{\mathrm{bgr}}$}
\KwOut{Enhanced RGB image $\hat I_{\mathrm{rgb}}$}
\BlankLine
\textbf{(Optional) Illumination correction:} convert to gray, estimate background via morphological opening (kernel $\approx 2\%$ of min dimension), replace LAB luminance with normalized gray.\;
\textbf{CLAHE on luminance:} apply CLAHE (clip $=2.0$, tiles $=8\times 8$) on $L$ in LAB space.\;
\textbf{Unsharp mask:} Gaussian blur ($\sigma{=}1.0$), then $I \leftarrow (1{+}\alpha)I - \alpha\,\mathrm{blur}$ with $\alpha{=}0.5$.\;
\textbf{(Optional) Denoise:} mild NLM with \texttt{denoise\_h} if set.\;
Convert to RGB to obtain $\hat I_{\mathrm{rgb}}$; then apply training-set normalization (Eq.~above).\;
\caption{Document enhancement pipeline for PubTabNet scans/prints.}
\end{algorithm}
\noindent
This preprocessing consistently improves text legibility and boundary contrast on low-quality scans without introducing artifacts, and we found it unnecessary for high-quality born-digital sources. For reproducibility, we keep parameters fixed across the corpus and \emph{never} compute normalization statistics on validation/test.

\subsubsection*{Structure-Head Targets and Key-Bias}
\label{sec:struct-head}

\paragraph{Structure-head targets}
The structure head predicts three dense maps capturing table geometry: horizontal separators (\emph{rows}), vertical separators (\emph{cols}), and \emph{corners}. Given a page image size $(H_{\!img},W_{\!img})$ and the ground-truth \texttt{HTML} with per-cell coordinates encoded as tokens \texttt{<x\_k>}, \texttt{<y\_k>}, we build supervision in three steps:

\emph{(i) Owner grid from HTML.}
We parse the table DOM and expand \texttt{rowspan}/\texttt{colspan} to form an integer grid $O\!\in\!\mathbb{Z}^{R\times C}$ where each entry stores the \emph{HTML-cell id} that occupies that slot (identical ids across a span). In parallel we recover a map $\text{bbox}:\{\text{cell id}\}\!\to\!(x_1,y_1,x_2,y_2)$ by reading the first/last coordinate markers in each cell’s inner HTML.

\emph{(ii) Boundary alignment.}
True row/column boundaries occur where neighboring owners differ:
\[
H_{\text{sep}}(r,c)=\mathbb{1}[O_{r,c}\neq O_{r+1,c}],
\]
\[
V_{\text{sep}}(r,c)=\mathbb{1}[O_{r,c}\neq O_{r,c+1}].
\]
We initialize boundary positions uniformly, then \emph{snap} them to available evidence from cell boxes: for a row boundary $r$, collect candidates from the bottom $y$ of cells above and the top $y$ of cells below; take the median to set the boundary $y$. Columns are treated analogously with $x$-coordinates. Monotone clipping enforces valid, non-crossing boundaries within $[0,H_{\!img}\!-\!1]$ and $[0,W_{\!img}\!-\!1]$.

\emph{(iii) Rasterization.}
From the aligned boundary positions, we draw soft separator fields by placing 1D Gaussian \emph{ridges} along each boundary segment:
\[
\text{RowMap}(y,x)\;=\;\max_{k}\exp\!\Big(-\tfrac{(y-y_k)^2}{2\sigma_{\ell}^2}\Big),
\]
\[
\text{ColMap}(y,x)\;=\;\max_{m}\exp\!\Big(-\tfrac{(x-x_m)^2}{2\sigma_{\ell}^2}\Big),
\]
restricted to valid spans between adjacent columns and rows. Corners are obtained as a smoothed intersection of row/col ridges (Gaussian blur with $\sigma_c$). Stacking yields the target tensor
\[
T=\big[\text{RowMap},\ \text{ColMap},\ \text{CornerMap}\big]\in[0,1]^{3\times H_{\!img}\times W_{\!img}}.
\]
At training time we downsample or predict at the structure-head resolution $(H_s,W_s)$ and use a standard pointwise loss (e.g., BCE with logits) against $T$.
\paragraph{Key-bias for cross-attention.}
At inference, structure maps modulate decoder cross-attention by adding a \emph{key-only bias} to encoder tokens, encouraging the decoder to attend near predicted separators and cell corners.

Let $S\!\in\!\mathbb{R}^{B\times 3\times H_s\times W_s}$ be structure-head logits (channels: rows/cols/corners). We convert to probabilities $P=\sigma(S)$ and bilinearly resize each channel to the encoder grid $(H_f,W_f)$:
\[
P_r,P_c,P_{\text{cor}}\ \in\ [0,1]^{B\times H_f\times W_f}.
\]
We derive axis profiles by max-pooling across the orthogonal axis,
\[
R_b(y)=\max_x P_r(y,x),\qquad C_b(x)=\max_y P_c(y,x),
\]
and broadcast back to 2D, then combine with the corners channel:
\[
B\;=\;\alpha\,R+\beta\,C+\gamma\,P_{\text{cor}}\ \in\ \mathbb{R}^{B\times H_f\times W_f}.
\]
To keep the attention softmax numerically stable, we apply a per-sample z-score to $B$ (flattened over space), and scale by a confidence factor derived from the mean binary entropy of $(P_r,P_c,P_{\text{cor}})$:
\begin{equation}
\begin{aligned}
\text{conf}
&= 1 - \frac{1}{3H_fW_f}\sum_{y,x}\!\Big[H(P_r)+H(P_c)+H(P_{\text{cor}})\Big],\\
\hat B
&= \lambda_0\,\text{conf}\cdot \mathrm{zscore}(B).
\end{aligned}
\end{equation}

Finally we clamp $\hat B$ to $[-c,c]$ and (optionally) stop gradients. Flattening over space gives $b_{\text{key}}\!\in\!\mathbb{R}^{B\times(H_fW_f)}$, which is \emph{added} to the cross-attention logits along the key dimension before the softmax. Intuitively, rows/columns contribute broad bands, corners sharpen locality, z-scoring equalizes samples, and entropy scaling down-weights uncertain maps. The implementation mirrors this formula with tunable weights $(\alpha,\beta,\gamma)$ and scale $\lambda_0$.

\begin{figure}[t]
  \centering
  \includegraphics[width=\columnwidth]{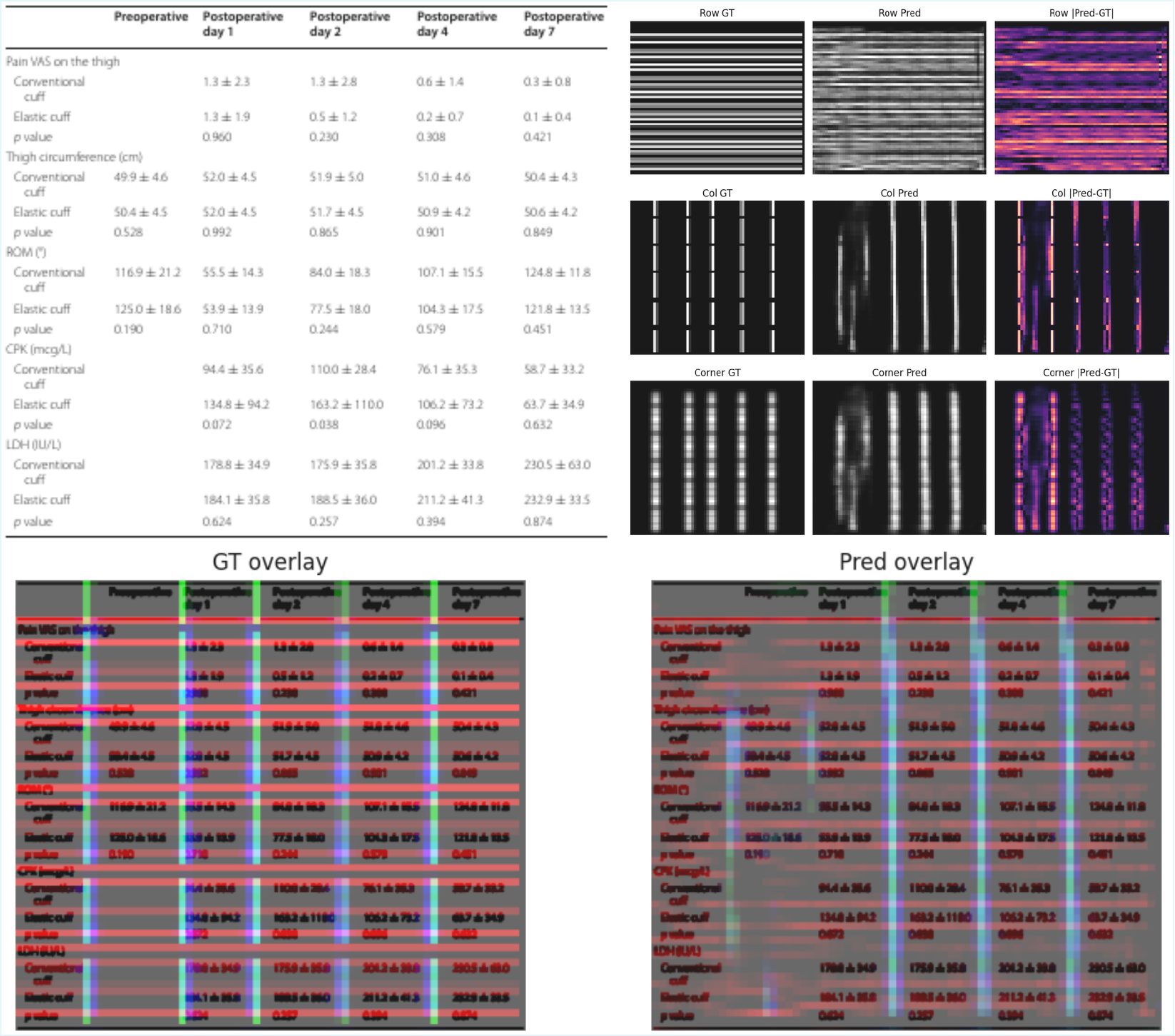}
  \caption{PubTabNet example with predicted priors (rows/cols/corners) and supervision targets from the structure head.}
  \label{fig:structure_samples}
\end{figure}

\section{Experiments}

\paragraph{Training setup}
We train a separate model for each dataset on a single NVIDIA H200 GPU (140\,GB). Batch sizes are 12, 12, 10, and 12 for PubTabNet, FinTabNet, SciTSR, and PubTables-1M, respectively. For SciTSR, table images are rasterized from source PDFs at 150\,dpi. On PubTabNet, FinTabNet, and SciTSR, we adopt a simple synthetic-to-real curriculum: training starts with 100\% synthetic tables, and the synthetic ratio is linearly annealed to 0\% by the end of training. For PubTables-1M, however, we do not use synthetic data. All models are trained with teacher forcing and 3\% error injection, and optimized with Adam using a learning-rate schedule decaying from $5\times10^{-5}$ to $5\times10^{-7}$.

\paragraph{Datasets}
We evaluate on four public benchmarks. For end-to-end TSR/TCR, we use PubTabNet~\cite{zhong2020image}, FinTabNet~\cite{zheng2021global}, and SciTSR~\cite{chi2019complicated}. PubTabNet provides about $500$k/$9$k/$9$k train/validation/test table images with \texttt{HTML} structure annotations, cell text, and cell boxes for the training and validation splits; the official test split does not include cell boxes. FinTabNet contains tables extracted from financial reports, annotated with \texttt{HTML} structure, cell text, and cell boxes. SciTSR focuses on scientific tables with complex structures such as spanning cells and hierarchical headers; we adopt its standard image-based evaluation protocol. To broaden the evaluation beyond TEDS/CAR-based benchmarks, we also report TSR results on PubTables-1M~\cite{Smock2022PubTables1M}, a large-scale benchmark derived from scientific articles and evaluated with the GriTS family of metrics.

To assess cross-dataset generalization, we further report \emph{zero-shot} transfer to ICDAR~2013: models are trained on \emph{either} SciTSR or FinTabNet and evaluated on ICDAR~2013 without adaptation. All benchmarks contain both simple and complex tables, including spans, nested headers, and grouped columns. To improve robustness on complex layouts, we generate additional synthetic training samples for PubTabNet, FinTabNet, and SciTSR through structure-aware augmentation, including span insertion/expansion, header nesting, and column grouping, with randomized rendered cell content. No synthetic data are used for PubTables-1M.

\subsection{Table Structure and Content Recognition}

For \textbf{PubTabNet} and \textbf{FinTabNet}, we evaluate end-to-end table structure and content recognition using TEDS and S\mbox{-}TEDS. Predicted tables are parsed as \texttt{HTML} DOM trees; TEDS measures tree-edit similarity with a leaf-level text edit term, while S\mbox{-}TEDS removes the text term to assess structure alone. For \textbf{SciTSR}, we follow the official Cell Adjacency Relations (CAR) protocol, which evaluates structural recovery through cell-adjacency graphs rather than text content: predicted and ground-truth cells are matched at $\mathrm{IoU}_{50}$, horizontal and vertical adjacency pairs are formed, and precision, recall, and F1 are computed against the reference graph. Predictions produced as \texttt{HTML} are deterministically converted to the relational format required by CAR. For \textbf{ICDAR~2013}, we adopt the same CAR protocol to isolate structural fidelity under zero-shot transfer.

\begin{table*}[t]
  \centering
  \small
  \caption{Comparison on PubTabNet, FinTabNet, and SciTSR. PubTabNet and FinTabNet are reported with TEDS/S-TEDS, while SciTSR is reported with CAR precision/recall/F1. Methods that require pre-supplied text regions or heavy external OCR/post-processing are excluded (e.g., TFLOP~\cite{khang2025tflop}). Best and second-best scores are highlighted in bold and underlined.}
  \label{tab:tsr_tcr_results}
  \setlength{\tabcolsep}{6pt}
  \begin{tabular}{p{0.14\textwidth} p{0.16\textwidth} cc cc ccc}
    \toprule
    \textbf{Family} & \textbf{Method} &
    \multicolumn{2}{c}{\textbf{PubTabNet}} &
    \multicolumn{2}{c}{\textbf{FinTabNet}} &
    \multicolumn{3}{c}{\textbf{SciTSR}}\\
    \cmidrule(lr){3-4}\cmidrule(lr){5-6}\cmidrule(lr){7-9}
    & & \textbf{TEDS} & \textbf{S-TEDS} & \textbf{TEDS} & \textbf{S-TEDS} & \textbf{P} & \textbf{R} & \textbf{F1}\\
    \midrule
    \multirow{6}{*}{\textbf{Split-and-Merge}}
      & TSRFormer~\cite{lin2022tsrformer}            & -- & \best{97.50} & -- & 98.40 & 99.70 & \best{99.60} & \secondbest{99.60}\\
      & TRUST~\cite{guo2208trust}                    & 96.20 & 97.10 & -- & -- & -- & -- & --\\
      & SEMv2~\cite{zhang2023semv2}                  & -- & \best{97.50} & -- & --  & -- & -- & --\\
      & SEM~\cite{zhang2022split}                    & -- & 96.30 & -- & -- & 97.70 & 96.52 & 97.11\\
      & SPLERGE~\cite{tensmeyer2019deep}             & \secondbest{96.27} & -- & -- & -- & 96.69 & 94.15 & 95.40\\
      & TABLET~\cite{hou2025tablet}                  & -- & -- & \best{98.54} & \best{98.71} & -- & -- & --\\
    \midrule
    \multirow{3}{*}{\textbf{Bottom-Up}}
      & LGPMA~\cite{Qiao2021LGPMA}                   & 94.60 & 96.70 & -- & -- & 98.20 & 99.30 & 98.80 \\
      & NCGM~\cite{Liu2022NCGM}                      & -- & 95.40 & -- & -- & 99.70 & \best{99.60} & \secondbest{99.60}\\
      & GTE~\cite{zheng2021global}                   & -- & 93.0 & -- & 91.0 & -- & -- & --\\
    \midrule
    \multirow{6}{*}{\textbf{Img-to-Seq}}
      & EDD~\cite{zhong2020image}                    & 88.30 & 89.90 & -- & 90.60 & -- & -- & --\\
      & Omniparser~\cite{wan2024omniparser}          & 88.83 & 90.45 & 89.75 & 91.55 & -- & -- & --\\
      & TableFormer~\cite{nassar2022tableformer}     & 93.60 & 96.75 & -- & 96.80 & -- & -- & --\\
      & VAST~\cite{huang2023improving}               & \best{96.31} & \secondbest{97.23} & 98.21 & 98.63 & \secondbest{99.77} & 99.26 & 99.51\\
      & UniTabNet~\cite{zhang2024unitabnet}          & -- & \best{97.50} & -- & -- & -- & -- & --\\
      & TableSeq                                     & 95.23 & 96.83 & 97.45 & \secondbest{98.69} & \best{99.79} & \secondbest{99.54} & \best{99.66}\\
    \bottomrule
  \end{tabular}
\end{table*}

Table~\ref{tab:tsr_tcr_results} reports only end-to-end systems that consume the table image as the sole input, grouped by inference paradigm (Split\mbox{-}and\mbox{-}Merge, Bottom-Up, Image-to-Sequence). On \textbf{SciTSR}, TableSeq attains the highest precision and F1, together with the second-highest recall among listed methods. On \textbf{FinTabNet}, it achieves the second-best S\mbox{-}TEDS overall (behind TABLET) and the best S\mbox{-}TEDS within the Image-to-Sequence family. On \textbf{PubTabNet}, it remains competitive on both TEDS and S\mbox{-}TEDS.

\paragraph{Evaluation on PubTables-1M}
To complement the TEDS/CAR-based benchmarks above, we also assess TableSeq on \textbf{PubTables-1M}~\cite{Smock2022PubTables1M} using the GriTS metrics, which measure grid-level similarity in terms of topology (GriTS$_\text{Top}$), content (GriTS$_\text{Con}$), and localization (GriTS$_\text{Loc}$). As shown in Table~\ref{tab:pubtables_results}, TableSeq achieves \textbf{99.10} GriTS$_\text{Top}$, \textbf{98.82} GriTS$_\text{Con}$, and \textbf{95.63} GriTS$_\text{Loc}$. It ranks second on GriTS$_\text{Top}$ and GriTS$_\text{Con}$ behind VAST, and second on GriTS$_\text{Loc}$ behind DETR while improving over VAST by \(+0.64\) on localization. These results indicate that the proposed single-stream formulation remains effective under the PubTables-1M evaluation protocol and generalizes well to a large-scale born-digital benchmark. Representative qualitative predictions and failure cases are provided in Appendix~\ref{app:qual_samples}.

\begin{table}[t]
  \centering
  \small
  \caption{TSR comparison on \textbf{PubTables-1M} (test). Metrics are GriTS$_\text{Top}$, GriTS$_\text{Con}$, and GriTS$_\text{Loc}$ (\%; higher is better).}
  \label{tab:pubtables_results}
  \begingroup
  \fontsize{8}{10.5}\selectfont
  \setlength{\tabcolsep}{4pt}
  \begin{tabular}{lccc}
    \toprule
    \textbf{Method} & \textbf{Top} & \textbf{Con} & \textbf{Loc} \\
    \midrule
    Faster R-CNN~\cite{Smock2022PubTables1M} & 86.16 & 85.38 & 72.11 \\
    DETR~\cite{Smock2022PubTables1M}         & 98.45 & 98.46 & \best{97.81} \\
    VAST~\cite{huang2023improving}           & \best{99.22} & \best{99.14} & 94.99 \\
    \midrule
    TableSeq                                 & \secondbest{99.10} & \secondbest{98.82} & \secondbest{95.63} \\
    \bottomrule
  \end{tabular}
  \endgroup
\end{table}

To assess cell content bounding-box detection, we compare against state-of-the-art detectors on the PubTabNet validation set (Table~\ref{tab:bbox_detection}). TableSeq achieves the highest AP$_{50}$, slightly outperforming TabGuard while remaining fully end-to-end.

\begin{table}[t]
  \centering
  \small
  \caption{Cell content bounding-box detection on \textbf{PubTabNet} (validation). Metric is AP$_{50}$ (\%).}
  \label{tab:bbox_detection}
  \begingroup
  \fontsize{9}{10.5}\selectfont
  \begin{tabularx}{0.55\linewidth}{l c}
    \toprule
    \multicolumn{2}{c}{PubTabNet} \\
    \midrule
    \textbf{Method} & \textbf{AP$_{50}$} \\
    \midrule
    EDD + BBox~\cite{nassar2022tableformer}  & 79.2 \\
    TableFormer~\cite{nassar2022tableformer} & 82.1 \\
    VAST~\cite{huang2023improving}           & 94.8 \\
    TabGuard~\cite{raja2025treading}         & 96.2 \\
    TableSeq                                 & \textbf{96.5} \\
    \bottomrule
  \end{tabularx}
  \endgroup
\end{table}

For cross-dataset generalization, we perform \emph{zero-shot} transfer to ICDAR~2013. Models are trained on either SciTSR or FinTabNet and evaluated directly on ICDAR~2013 without adaptation. Following the SciTSR setup, \texttt{HTML} predictions are converted to cell adjacency relations and scored with CAR at $\mathrm{IoU}_{50}$, isolating structure independent of text recognition and dataset-specific heuristics. Results are summarized in Table~\ref{tab:icdar13-xfer}. \begin{table}[t] \centering \small \caption{Zero-shot transfer to \textbf{ICDAR~2013} (structure only). Each model is trained on a single source dataset and evaluated on ICDAR~2013 without adaptation. Metrics follow the CAR protocol (P/R/F1, \%).} \label{tab:icdar13-xfer} \begingroup \fontsize{8}{10.5}\selectfont \begin{tabularx}{\linewidth}{l c c c c} \toprule \multicolumn{5}{c}{ICDAR~2013} \\ \midrule \textbf{Method} & \textbf{Training Dataset} & \textbf{P$\uparrow$} & \textbf{R$\uparrow$} & \textbf{F1$\uparrow$} \\ \midrule LGPMA~\cite{Qiao2021LGPMA} & SciTSR & 93.00 & \secondbest{97.70} & 95.30 \\ GTE~\cite{zheng2021global} & FinTabNet & 92.72 & 94.41 & 93.50 \\ VAST~\cite{huang2023improving} & SciTSR & 93.84 & 97.68 & \secondbest{95.72} \\ VAST~\cite{huang2023improving} & FinTabNet & \secondbest{95.29} & \best{97.79} & 96.52 \\ TableSeq & SciTSR & \best{95.78} & 97.53 & \best{96.60} \\ \bottomrule \end{tabularx} \endgroup \end{table}

These results establish that TableSeq is competitive not only on in-domain benchmarks, but also under cross-dataset transfer. We next examine two complementary aspects of the proposed sequence formulation. First, because the model remains autoregressive, we study whether decoding can be accelerated without modifying the backbone or sacrificing much accuracy. Second, since TableSeq generates a unified representation of structure and content, we investigate whether the same interface can support structured, index-based table querying beyond conventional TSR/TCR evaluation.

\subsection{Multi-token Prediction (MTP)} Autoregressive decoding makes latency scale with sequence length, which is nontrivial for image-to-sequence TSR/TCR because the output interleaves \texttt{HTML} tags, text tokens, and discretized coordinate tokens. To reduce latency without altering the backbone, we employ multi-token prediction (MTP) inspired by recent LLM work~\cite{gloeckle2404better}: at decoding step $t$, a single forward pass predicts the next $n$ tokens and inference proceeds blockwise so that each outer step emits up to $n$ tokens before advancing. Training minimizes a weighted sum of token-level cross-entropies under teacher forcing, \begin{equation} \mathcal{L}_{\text{mtp}} = \sum_{i=1}^{n} w_i \,\mathrm{CE}\!\left(\hat{\mathbf{y}}^{(i)}_{t+i},\, \mathbf{y}_{t+i}\right), \qquad \sum_{i=1}^{n} w_i = 1. \end{equation} On \textbf{PubTabNet} (test), the baseline configuration ($n{=}1$) runs at a mean wall-clock latency of \textbf{1.80\,s/img}; see Table~\ref{tab:multitokens-results}. Increasing the block size yields near-linear speedups with small accuracy trade-offs: at $n{=}2$ latency drops to \textbf{1.25\,s/img} ($\approx$31\% faster) with \mbox{S-TEDS}/TEDS changing by $-1.01$/$-0.88$ points; at $n{=}4$ latency reaches \textbf{0.95\,s/img} ($\approx$47\% faster) with \mbox{S-TEDS}/TEDS changes of $-1.36$/$-1.11$ points. In practice, $n{=}2$ offers a balanced operating point, while $n{=}4$ is preferable for high-throughput settings where a modest accuracy decrease is acceptable. \begin{table}[t] \centering \small \caption{MTP ablation on \textbf{PubTabNet} (test). Metrics are \% (↑ better). Latency is mean wall-clock seconds per image (↓ better), measured on a single NVIDIA H200 (140\,GB), batch size 1.} \label{tab:multitokens-results} \begingroup \fontsize{8}{10.5}\selectfont \begin{tabularx}{\linewidth}{l c c c c} \toprule \textbf{Method} & \textbf{Tok/step} & \textbf{S-TEDS$\uparrow$} & \textbf{TEDS$\uparrow$} & \textbf{Latency$\downarrow$} \\ \midrule TableSeq & 1 & 96.83 & 95.23 & 1.80 \\ TableSeq & 2 & 95.82 & 94.35 & 1.25 \\ TableSeq & 4 & 95.47 & 94.12 & 0.95 \\ \bottomrule \end{tabularx} \endgroup \end{table}

\subsection{Hierarchical Recognition Tasks}
While the preceding experiments evaluate TableSeq as a TSR/TCR system, its single-stream output interface is not restricted to \texttt{HTML} reconstruction alone. If the model captures table structure and cell content in a coherent representation, it should also support structured querying over the recognized table without task-specific heads. To test this property, we evaluate index-based recognition beyond TSR/TCR following Zhou et al.~\cite{zhou2024enhancing}. \emph{Index-based Cell Recognition} (ICR) returns the content of the cell at $(i,j)$; \emph{Row Index-based Data Recognition} (IRDR) returns the left-to-right sequence in row $i$; and \emph{Column Index-based Data Recognition} (ICDR) returns the top-to-bottom sequence in column $j$. We adopt the NGTR protocol on \textbf{SciTSR} (official test split) and \textbf{PubTabNet} (a random 1{,}500-image subset of the validation set), reusing NGTR’s prompt templates. For ICR, the model outputs a single normalized string; for IRDR/ICDR, it returns a delimiter-separated list encoding the ordered cells. Although our model also produces bounding boxes, evaluation is text-only. We normalize whitespace, case, punctuation, and number formats; merged cells are assigned to their top-left grid coordinate, and row/column spans are expanded when constructing lists. Baselines include open-source \emph{Phi-3.5-Vision-Instruct} (Phi) and \emph{Llama-3.2-90B-Vision-Instruct} (Llama), and closed-source \emph{GPT-4o-mini} (GPT-mini), \emph{QwenVL-Max} (Qwen), \emph{GPT-4o} (GPT), and \emph{Gemini-1.5-Pro} (Gemini). Following~\cite{zhou2024enhancing}, we report ACC for ICR (exact match after normalization) and micro-F1 for IRDR/ICDR computed over tokenized cell lists. Table~\ref{tab:ngtr-index-vertical} reports results on the pooled evaluation set formed by combining the SciTSR test split and a 1,500-image PubTabNet validation subset. \begin{table}[t] \centering \small \caption{Index-style recognition (IRDR/ICDR/ICR) on the pooled \textbf{SciTSR} test split and a 1,500-image \textbf{PubTabNet} validation subset. Metrics are percentages.} \label{tab:ngtr-index-vertical} \begingroup \fontsize{8}{10.5}\selectfont \begin{tabularx}{\linewidth}{lccc} \toprule \textbf{Method} & \textbf{IRDR (F1$\uparrow$)} & \textbf{ICDR (F1$\uparrow$)} & \textbf{ICR (ACC$\uparrow$)} \\ \midrule GPT & \secondbest{52.85} & \secondbest{69.21} & 37.20 \\ GPT-mini & 30.32 & 54.78 & 16.20 \\ Gemini & 48.45 & \best{71.33} & \best{44.40} \\ Qwen & 30.32 & 55.20 & 23.20 \\ Llama & 36.56 & 54.63 & 21.61 \\ Phi & 19.48 & 37.49 & 7.20 \\ \midrule Ours & \best{67.73} & 67.50 & \secondbest{42.85} \\ \bottomrule \end{tabularx} \endgroup \end{table} 

Our method achieves the highest IRDR score (67.73), surpassing the strongest baseline (GPT at 52.85) by +14.88 points, indicating that row-index conditioning aligns well with left-to-right aggregation. On ICDR, our score (67.50) is competitive within 3.83 of the best system (Gemini at 71.33) and 1.71 behind GPT (69.21), suggesting that column-wise traversal remains slightly more challenging, likely due to vertically merged headers and multi-line cells. For ICR, our accuracy (42.85) ranks second, only 1.55 below Gemini (44.40) and +5.65 above GPT (37.20), reflecting reliable localization and reading for targeted cells. Overall, the simple index-conditioned interface yields state-of-the-art IRDR performance and near–state-of-the-art ICDR/ICR without task-specific heads, with residual errors concentrated in vertical traversals and merged-span edge cases.

\section{Ablation Studies}
\subsection{Encoder Architecture}
We ablate the \emph{encoder} on \textbf{FinTabNet} to isolate backbone effects while holding fixed the training data, schedule, augmentations, and the decoder from Sec.~\ref{method}. Our default backbone is a lightweight convolutional network (FCN-H16). For transformer-based alternatives under comparable parameter budgets, we evaluate a ViTDet-style encoder~\cite{li2022exploring} configured following the document-image recipe of Vary~\cite{wei2024vary}, previously effective in GOT-OCR~2.0~\cite{wei2024general} and a Pix2Struct-style ViT (6 layers)~\cite{lee2023pix2struct} that accommodates variable input image sizes. To maintain parity with the CNN, the ViTDet depth is reduced to six layers. Inputs are resized to $1024\times1024$; the ViTDet-style encoder produces a fixed-length sequence of $256$ tokens with $1024$-dimensional embeddings ($\mathbf{X}\in\mathbb{R}^{256\times1024}$), which we reshape to a 2D grid for downstream processing. The Pix2Struct-style variant scales each image isotropically to fit a \emph{maximum patch budget} and crops any residual border; token embeddings are $768$-dimensional and are projected to the decoder width through a thin linear bridge. We report results for patch budgets of 1024 and 2048. All encoders feed the same decoder and share identical optimization and augmentation settings. Results on the FinTabNet test set are reported in Table~\ref{tab:encoder_variants}.

The convolutional baseline attains the strongest TEDS despite the smallest footprint: FCN-H16 reaches \textbf{97.45} with \textbf{30M} parameters, exceeding the best transformer variant (Pix2Struct, budget~2048) by \textbf{+3.57} points and the ViTDet-style encoder by \textbf{+5.12} points, while using only $\sim\!40\%$ fewer parameters than the ViT alternatives (50--52M). Within the Pix2Struct family, increasing the patch budget from 1024 to 2048 yields a modest \textbf{+0.37} improvement (93.51$\rightarrow$93.88) at fixed depth, indicating diminishing returns from finer sampling without commensurate model capacity. Under a tight compute budget and with our decoder, high-resolution local features from a compact CNN offer a better accuracy–efficiency trade-off than shallow ViT backbones.
We stress that this comparison is restricted to a from-scratch setting with matched training conditions and comparable parameter budgets; it does not imply that FCN-H16 would necessarily remain superior to heavily pretrained ViT-based encoders.

\begin{table}[t]
  \centering
  \small
  \caption{TableSeq on the FinTabNet test set with different encoder backbones (decoder and training held fixed). TEDS is reported in \%, Params denotes millions of parameters.}
  \label{tab:encoder_variants}
  \begingroup
  \fontsize{9}{10.5}\selectfont
  \begin{tabularx}{\linewidth}{lcc}
    \toprule
    \textbf{Encoder} & \textbf{TEDS} & \textbf{Params (M)} \\
    \midrule
    ViTDet-style                    & 92.33 & 52 \\
    ViT–Pix2Struct (1024 patches)   & 93.51 & 50 \\
    ViT–Pix2Struct (2048 patches)   & 93.88 & 50 \\
    FCN-H16 (TableSeq)              & \textbf{97.45} &  \textbf{30} \\
    \bottomrule
  \end{tabularx}
  \endgroup
\end{table}

\subsection{Impact of Model Design and Synthetic Data}

We quantify the effect of architectural and data choices on \textbf{PubTabNet} by decomposing TEDS over \emph{simple} and \emph{complex} tables. The ablation toggles five components while keeping the decoder, optimization, and data pipeline fixed: vertical sampling at $H'/32$ vs.\ $H'/16$ (exactly one active per row), a single Transformer encoder layer inserted before cross-attention (\textit{Tr}), a \textit{structure head} (\textit{SH}), an optional \textit{key-bias} at inference derived from the structure stream (\textit{KB}), and the inclusion of \textit{synthetic data} (\textit{Syn}). Table~\ref{tab:synth_data_impact} reports absolute TEDS for the $H'/32$ baseline (first row) and for the final model (last row), with intermediate rows showing the incremental per-subset gain obtained when the newly added component is enabled.

The baseline attains 93.07/96.95 on simple/complex tables. Moving to $H'/16$ yields consistent gains (+0.15/\secondbest{+0.19}), indicating that finer vertical sampling particularly benefits complex layouts with row spans and stacked headers. Adding a pre–cross-attention Transformer layer provides a smaller but uniform boost (+0.12/+0.05). Enabling the structure head improves alignment (\secondbest{+0.21}/+0.17). Using synthetic data contributes the largest single increment (\best{+0.40}/\best{+0.33}). In aggregate, TEDS rises by \textbf{+0.97} on simple and \textbf{+0.86} on complex tables, reaching \textbf{94.04}/\textbf{97.81} for the full configuration.
The full configuration corresponds to the overall PubTabNet TEDS of \textbf{95.23} reported in Table~\ref{tab:tsr_tcr_results}; Table~\ref{tab:synth_data_impact} is used only to decompose the gains over the simple and complex subsets.

\begin{table}[t]
  \centering
  \small
  \caption{TEDS on \textbf{PubTabNet} (test). The first row gives the $H'/32$ baseline; intermediate rows report the incremental subset-specific gains obtained at each step as components are added sequentially; the last row shows absolute TEDS of the full model (“TableSeq”). Exactly one of H32/H16 is active per row; \cmark\ indicates the component is enabled.}
  \label{tab:synth_data_impact}
  \begingroup
  \setlength{\tabcolsep}{5pt}
  \fontsize{8}{10.5}\selectfont
  \begin{tabularx}{\linewidth}{c c c c c c >{\centering\arraybackslash}X >{\centering\arraybackslash}X}
    \toprule
    \textbf{Syn} & \textbf{SH} & \textbf{KB} & \textbf{Tr} & \textbf{H32} & \textbf{H16} & \textbf{Simple TEDS} & \textbf{Complex TEDS} \\
    \midrule
      &   &   &   & \cmark &             & 93.07 & 96.95 \\
      &   &   &   &        & \cmark       & +0.15 & \secondbest{+0.19} \\
      &   &   & \cmark &    & \cmark      & +0.12 & +0.05 \\
      & \cmark &   & \cmark & & \cmark    & \secondbest{+0.21} & +0.17 \\
      & \cmark & \cmark & \cmark & & \cmark & +0.09 & +0.12 \\
      \cmark & \cmark & \cmark & \cmark & & \cmark & \best{+0.40} & \best{+0.33} \\
    \midrule
    \multicolumn{6}{r}{\textbf{TableSeq}} & \best{94.04} & \best{97.81} \\
    \bottomrule
  \end{tabularx}
  \endgroup
  \vspace{0.25em}

  \footnotesize
  \emph{Abbreviations.} Syn: synthetic data; SH: structure head; KB: key-bias; Tr: Transformer encoder layer; H32/H16: vertical resolution $H'/\{32,16\}$.
\end{table}

\subsection{Impact of Grid Quantization}

The default TableSeq discretizes coordinates with \texttt{5\,px} bins. To characterize the speed–accuracy trade-off, we ablate the grid unit $u\!\in\!\{2,5,8\}$ on \textbf{PubTabNet} (validation) while keeping the architecture, training schedule, and decoder fixed. The model emits structure and cell boxes in a single sequence; we therefore report TEDS for structure/content fidelity and AP$_{50}$ (IoU $=0.50$) for box localization. Reducing $u$ enlarges the coordinate-token vocabulary and slightly lengthens the generated sequence, but introduces no additional parameters.

Results in Table~\ref{tab:quantization_ablate} show that moving from a coarse grid (8\,px) to \texttt{5\,px} yields a clear gain in AP$_{50}$ (\,+1.22) alongside a modest TEDS improvement (\,+0.25). Further refining to \texttt{2\,px} produces only marginal localization gains over \texttt{5\,px} (\,+0.07 AP$_{50}$) and negligible change in TEDS ($-0.03$), at the cost of higher decoding latency. We thus retain \texttt{5\,px} as the default operating point and treat \texttt{2\,px} as a high-precision option when localization is the bottleneck.

\begin{table}[t]
  \centering
  \small
  \caption{Effect of coordinate quantization on \textbf{PubTabNet} (validation). TEDS measures structure/content fidelity (\%); AP$_{50}$ is average precision at IoU $=0.50$ (\%).}
  \label{tab:quantization_ablate}
  \begingroup
  \fontsize{9}{10.5}\selectfont
  \setlength{\tabcolsep}{6pt}
  \begin{tabularx}{0.65\linewidth}{c | >{\centering\arraybackslash}X >{\centering\arraybackslash}X}
    \toprule
    \textbf{Pixels / Unit} & \textbf{TEDS$\uparrow$} & \textbf{AP$_{50}\uparrow$} \\
    \midrule
    2 & 96.25 & 95.72 \\
    5 & 96.28 & 95.65 \\
    8 & 96.03 & 94.43 \\
    \bottomrule
  \end{tabularx}
  \endgroup
\end{table}

\subsection{Sensitivity of Key-Biased Cross-Attention}
\label{subsec:keybias_sensitivity}

To assess the robustness of the proposed key-bias mechanism, we perform a small sensitivity study on \textbf{PubTabNet} validation using the final trained checkpoint, varying only the \emph{inference-time} bias parameters while keeping the model weights fixed. We evaluate the global scale $\lambda_0$ and the relative weights $(\alpha,\beta,\gamma)$ that combine the row, column, and corner signals into the key bias. This isolates the effect of the bias itself from training noise and makes the study inexpensive to reproduce.

Table~\ref{tab:keybias_sensitivity} shows that setting $\lambda_0=0$ reduces S-TEDS from \textbf{96.83} to \textbf{95.79}, confirming that the key-bias contributes positively. Moderate scaling is the most effective: performance improves from \textbf{96.22} at $\lambda_0=0.5$ to a peak of \textbf{96.83} at $\lambda_0=1.0$, then decreases to \textbf{96.35} at $\lambda_0=2.0$, suggesting that excessively strong biasing can oversuppress non-structural evidence.

When varying the corner-channel weight $\gamma$ while keeping $\alpha=\beta=1$ and $\lambda_0=1$, the best result is obtained with the balanced setting $(1,1,1)$. However, performance remains relatively stable for nearby choices, with S-TEDS values of \textbf{96.61}, \textbf{96.68}, and \textbf{96.50} for $\gamma=0.5$, $1.5$, and $2.0$, respectively. Removing the corner term entirely (\(\gamma=0\)) also preserves most of the gain (\textbf{96.53}), indicating that the separator cues already provide a strong prior, while the corner signal offers an additional but not overly brittle improvement. Overall, these results show that key-biased cross-attention yields consistent gains and does not rely on narrowly tuned hyperparameters.

\begin{table}[t]
  \centering
  \small
  \caption{Sensitivity of key-biased cross-attention on \textbf{PubTabNet} test set. All results use the same trained checkpoint; only inference-time bias parameters are varied.}
  \label{tab:keybias_sensitivity}
  \begin{tabular}{c c c c c}
    \toprule
    $\alpha$ & $\beta$ & $\gamma$ & $\lambda_0$ & \textbf{S-TEDS} \\
    \midrule
    1.0 & 1.0 & 1.0 & 0.0 & 95.79 \\
    1.0 & 1.0 & 1.0 & 0.5 & 96.22 \\
    1.0 & 1.0 & 1.0 & 1.0 & \textbf{96.83} \\
    1.0 & 1.0 & 1.0 & 2.0 & 96.35 \\
    \midrule
    1.0 & 1.0 & 0.0 & 1.0 & 96.53 \\
    1.0 & 1.0 & 0.5 & 1.0 & 96.61 \\
    1.0 & 1.0 & 1.5 & 1.0 & 96.68 \\
    1.0 & 1.0 & 2.0 & 1.0 & 96.50 \\
    \bottomrule
  \end{tabular}

\end{table}

\section{Limitations}
TableSeq remains autoregressive, so inference latency still grows with sequence length. Multi-token prediction reduces decoding iterations and improves speed, but does not remove this dependency entirely. The discretized coordinate formulation also entails a trade-off between simplicity and localization precision, and some fine-grained boundary errors may remain due to quantization.

Most residual errors occur in structurally ambiguous cases, including merged cells, grouped rows, weak separators, and column-wise traversal. Moreover, although the model shows some tolerance to mild rotations, it currently predicts rectangular boxes and is therefore not explicitly designed for strongly rotated or non-rectangular layouts. Addressing such cases would likely require richer geometric outputs, such as polygons, with additional decoding cost.

More broadly, the evaluation is constrained by the scope and quality of current public benchmarks, which may contain annotation ambiguities and do not fully reflect difficult real-world settings such as degraded scans, handwritten tables, multi-page documents, or low-resource scripts. Extending the evaluation to such conditions, and studying the effect of stronger pretrained visual encoders, are important directions for future work.

\section{Conclusion}
We introduced TableSeq, a minimal image-to-sequence framework that jointly predicts structure, content, and cell boxes. Despite its simplicity, the model is competitive across PubTabNet, FinTabNet, SciTSR, and PubTables-1M: it attains leading CAR scores on SciTSR, the best S-TEDS within the image-to-sequence family on FinTabNet, and strong performance on PubTabNet. Ablations show that a compact FCN-H16 encoder outperforms parameter-matched ViT variants, and that lightweight additions (a single transformer layer, a structure head, and a simple synthetic-data curriculum) yield consistent gains. Beyond TSR/TCR, the same interface transfers to index-based recognition tasks, achieving the best IRDR and near-best ICDR/ICR. Future work will target faster decoding and stronger vertical reasoning (e.g., span-consistent constraints), while extending evaluation to more diverse document settings.

\section*{Acknowledgements}
This work was granted access to the HPC resources of CRIANN (Regional HPC Center, Normandy, France) and GENCI-IDRIS. The authors gratefully acknowledge this computational support.

\appendix
\section{Synthetic Data}
\label{app:synthetic_data}
\paragraph{High-level procedure}
We denote the page image by $I$, the HTML by $H$ (with per-cell coordinate tags \texttt{<x\_i>}, \texttt{<y\_j>}), and the scale factor from HTML units to image pixels by $s$.


\begin{algorithm}[t]
\small
\SetAlgoLined
\DontPrintSemicolon
\KwIn{Image $I$, HTML $H$ (per-cell coords), scale $s$, font set $\mathcal{F}$}
\KwOut{Edited image $\tilde I$, updated HTML $\tilde H$}
\BlankLine
\textbf{Parse cells.} $\mathcal{C}\!\leftarrow\!\{(b_i,t_i)\}$ from $H$, with pixel boxes via \textsc{NormBox}$(\cdot,s)$\tcp*[r]{\texttt{\_extract\_box\_from\_cell\_html}, \texttt{\_extract\_text}}
\BlankLine
\textbf{(Optional) structure augmentation.} With prob.\ $\{p_k\}$ apply a random subset:\;
\Indp
span merges/splits; header nesting; column grouping; limited row/col insert/delete; small layout jitter.\;
\Indm
Update $H$ and recompute boxes if changed to get $(H',\mathcal{C}')$; else set $(H',\mathcal{C}')\!\leftarrow\!(H,\mathcal{C})$.\;
\BlankLine
\ForEach{$(b_i,t_i)\in\mathcal{C}'$}{
  $\hat c_{\text{bg}}\!\leftarrow\!\textsc{BgColor}(I,b_i)$;\quad $(t,r,b,l)\!\leftarrow\!\textsc{EdgeThk}(I,b_i,\hat c_{\text{bg}})$\;
  $b_i^{\text{in}}\!\leftarrow\!\textsc{InnerRegion}(b_i,t,r,b,l)$ \tcp*[r]{safe region inside borders}
  Sample $f\!\sim\!\mathcal{F}$, choose alignments;\ $\tilde t_i\!\leftarrow\!\textsc{Synthesize}(t_i)$;\ $\tilde t_i\!\leftarrow\!\textsc{ClipCap}(\tilde t_i,b_i^{\text{in}})$\;
  $s^\star\!\leftarrow\!\textsc{FitFont}(\tilde t_i,f,b_i^{\text{in}})$;\quad \uIf{\textsc{Overflow}}{$\tilde t_i\!\leftarrow\!\textsc{Truncate}(\tilde t_i)$; $s^\star\!\leftarrow\!\textsc{FitFont}(\tilde t_i,f,b_i^{\text{in}})$}
  \textsc{Wipe}$(I,b_i^{\text{in}},\hat c_{\text{bg}})$;\quad $\hat b_i\!\leftarrow\!\textsc{Render}(I,\tilde t_i,f,s^\star,\text{align})$\;
  $\bar b_i\!\leftarrow\!\textsc{Normalize}(\hat b_i,s^{-1})$;\quad update inner HTML of $H'$ with coords $(\bar b_i)$ and escaped $\tilde t_i$.\;
}
\Return $\tilde I\!\leftarrow\! I,\ \tilde H\!\leftarrow\! H'$.\;
\caption{Synthetic page generation by in-place editing with optional structure augmentation.}
\end{algorithm}

\section{Qualitative Prediction Samples and Failure Cases}
\label{app:qual_samples}

We provide representative qualitative examples to complement the quantitative results reported in the main paper. Each sample shows the input table image together with the rendered ground-truth HTML and the rendered prediction. These examples illustrate both the strengths of TableSeq and its remaining failure modes, especially for grouped rows, empty leading cells, and ambiguous span configurations.

\begin{figure*}[p]
  \centering
  \includegraphics[width=0.96\textwidth]{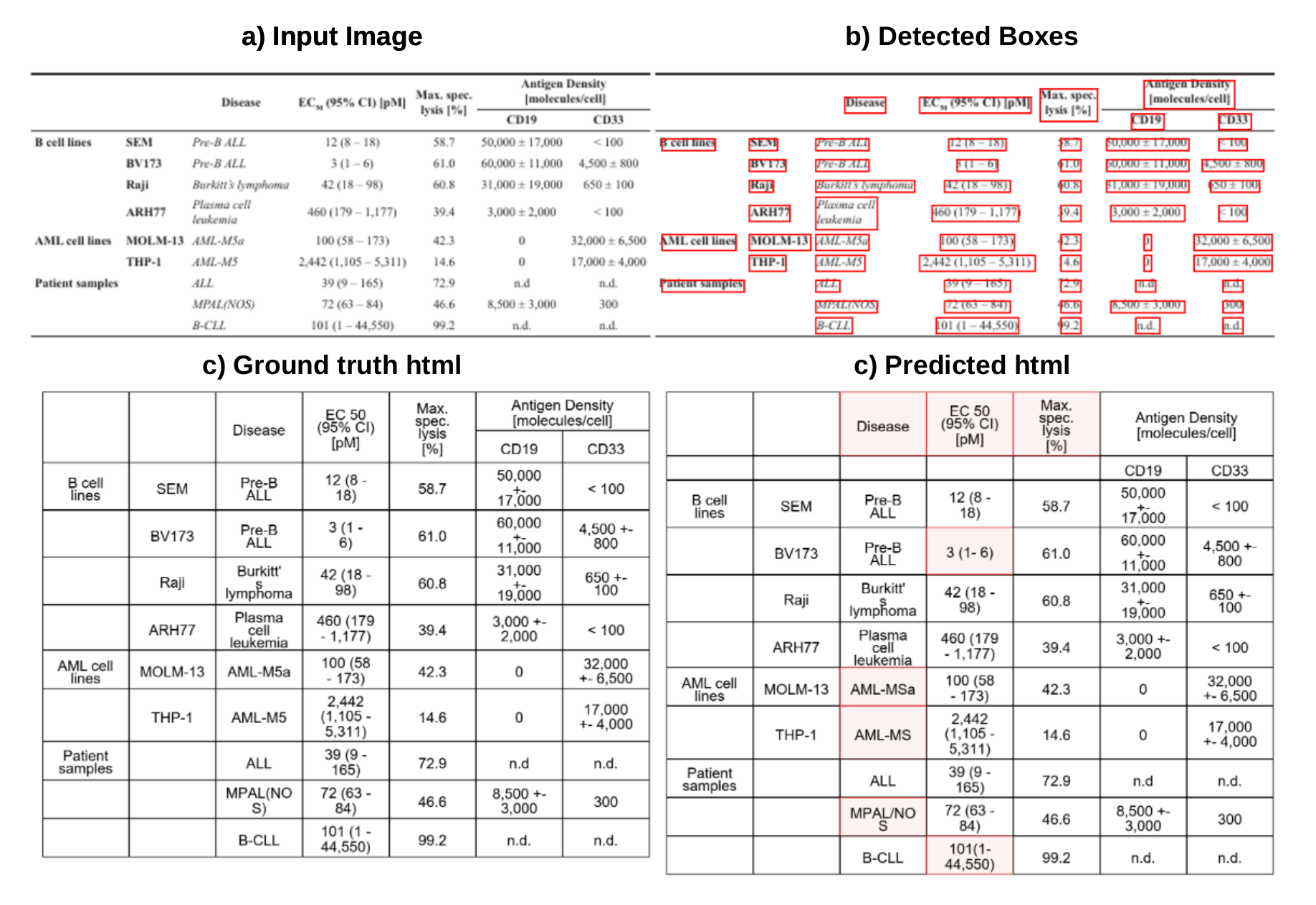}
  \caption{Representative prediction on a dense scientific table. TableSeq correctly recovers most rows, columns, and cell contents, while residual errors are concentrated around empty leading cells and the left grouping structure.}
  \label{fig:qual_ptn}
\end{figure*}

\begin{figure*}[p]
  \centering
  \includegraphics[width=0.96\textwidth]{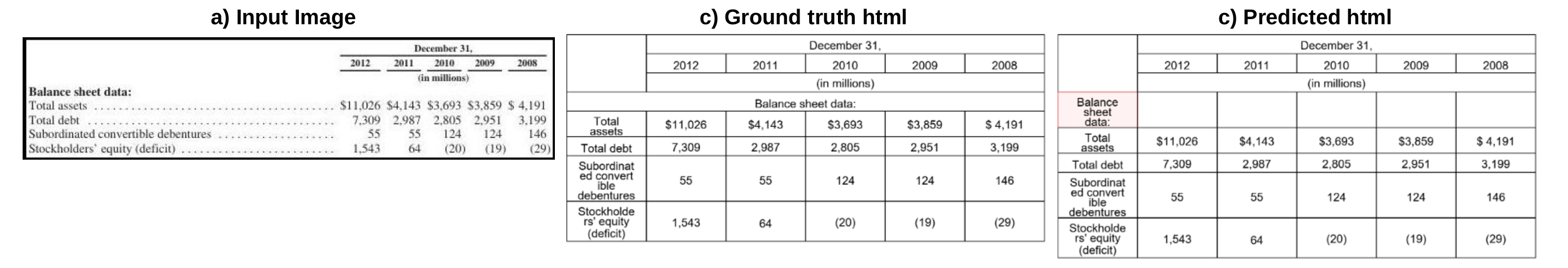}
  \caption{Representative prediction on a financial table from FinTabNet. Most headers and numerical values are recovered correctly, but the left stub column and blank header region remain challenging.}
  \label{fig:qual_ftn}
\end{figure*}

\begin{figure*}[p]
  \centering
  \includegraphics[width=0.96\textwidth]{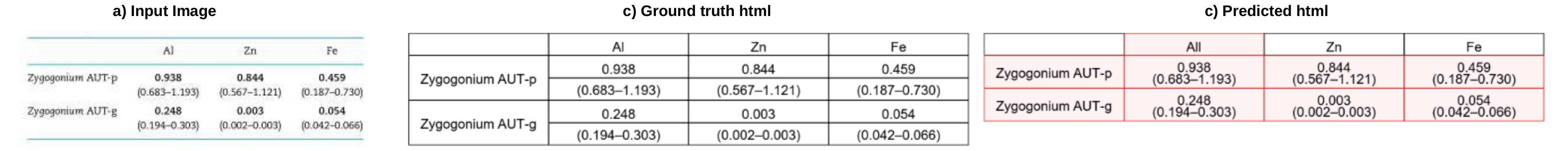}
  \caption{Representative prediction on a grouped-row table. The global layout is largely preserved, but the first-column grouping and row-span assignment remain ambiguous.}
  \label{fig:qual_pubtables}
\end{figure*}

\end{document}